%% file: main.tex
\documentclass[runningheads]{llncs}
\usepackage{graphicx}
\usepackage{comment}
\usepackage{amsmath,amssymb} 
\usepackage{color}
\usepackage{multirow}
\usepackage{booktabs}
\usepackage{subcaption}
\usepackage{float}
\usepackage[normalem]{ulem}
\usepackage{textcomp}
\usepackage{enumitem}

\begin{document}
\pagestyle{headings}
\mainmatter
\def\ECCVSubNumber{6790}  

\title{\textit{The Liar's Walk} \\ Detecting Deception with Gait and Gesture} 


\titlerunning{The Liar's Walk}
%
\author{Tanmay Randhavane\inst{1} \and
Uttaran Bhattacharya\inst{2} \and
Kyra Kapsaskis\inst{1} \and 
Kurt Gray\inst{1} \and
Aniket Bera\inst{2} \and
Dinesh Manocha\inst{2}
}
\authorrunning{T. Randhavane et al.}
%
\institute{University of North Carolina, Chapel Hill, NC 27516, USA
\email{tanmay@cs.unc.edu,\{kyrakaps,kurtjgray\}@gmail.com}\and
University of Maryland, College Park, MD 20742, USA \\
\email{\{uttaranb,ab,dm\}@cs.umd.edu}\\
\url{http://gamma.cs.unc.edu/GAIT/}} 

\newcommand{\accuracy}{$88.41\%$}
\newcommand{\accuracybold}{$\mathbf{88.41\%}$}
\newcommand\Tstrut{\rule{0pt}{2.6ex}} 
\newcommand\Bstrut{\rule[-0.9ex]{0pt}{0pt}}

\maketitle

\begin{abstract}
We present a data-driven deep neural algorithm for detecting deceptive walking behavior using nonverbal cues like gaits and gestures. We conducted an elaborate user study, where we recorded many participants performing tasks involving deceptive walking. We extract the participants' walking gaits as series of 3D poses. We annotate various gestures performed by participants during their tasks. Based on the gait and gesture data, we train an LSTM-based deep neural network to obtain deep features. Finally, we use a combination of psychology-based gait, gesture, and deep features to detect deceptive walking with an accuracy of \accuracy. This is an improvement of $10.6\%$ over handcrafted gait and gesture features and an improvement of $4.7\%$ and $9.2\%$ over classifiers based on the state-of-the-art emotion and action classification algorithms, respectively. Additionally, we present a novel dataset, \textit{DeceptiveWalk}, that contains gaits and gestures with their associated deception labels. To the best of our knowledge, ours is the first algorithm to detect deceptive behavior using non-verbal cues of gait and gesture.
\end{abstract}

\input{1_introduction}
\input{2_related}
\input{3_approach}

\input{4_classification}
\input{5_results}
\input{6_conclusion}

%
%
\bibliographystyle{splncs04}
\bibliography{template}
\end{document}

%% file: 1_introduction.tex
\section{Introduction}
\begin{figure}[t]
      \centering
      \includegraphics[width=\columnwidth]{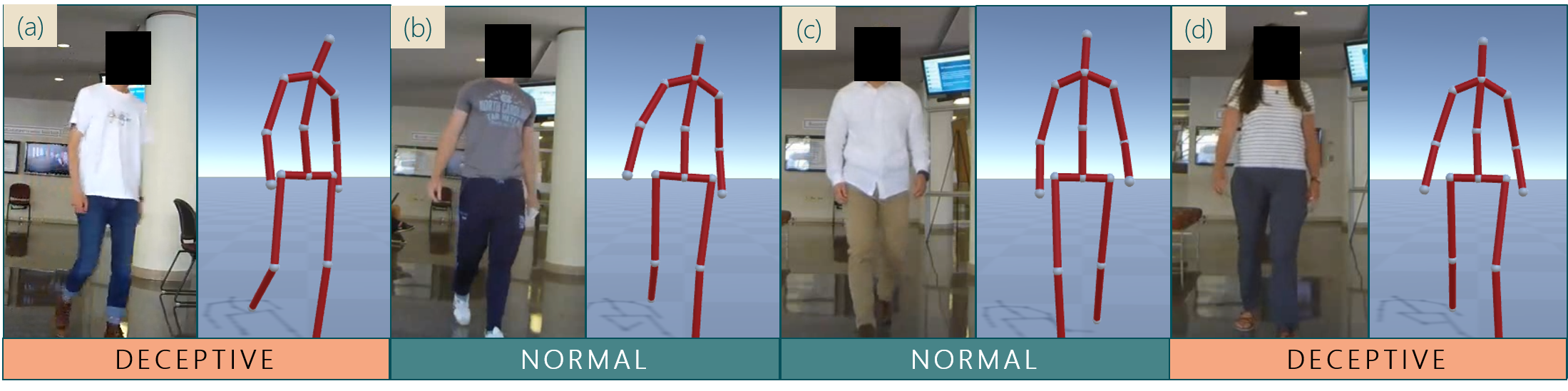}
      \caption{\textbf{Detecting Deception}: We present a new data-driven algorithm for detecting deceptive walking using nonverbal cues of gaits and gestures. We take an individual's walking video as an input, compute psychology-based gait features, gesture features, and deep features learned from a neural network, and combine them to detect deceptive walking. In the examples shown above, smaller hand movements (a) and the velocity of hands and feet joints (d) provide deception cues.}
      \vspace{-15pt}
      \label{fig:cover}
\end{figure}


In recent years, AI and vision communities have focused a lot on learning human behaviors, and human-centric video analysis has rapidly developed~\cite{wang2015video,wei2017deep,xu2016heterogeneous,mcduff2017large}. While conventional video content analysis pays attention to the analysis of the video content, human-centric video analysis focuses on the humans in the videos and attempts to obtain information about their behaviors, describe their dispositions, and predict their intentions. This includes recognition of emotions~\cite{kuo2018compact,lee2019context,marinoiu20183d}, personalities~\cite{wei2017deep,zhang2016deep}, and actions~\cite{wu2017recent,yao2019review}, as well as anomaly detection~\cite{nguyen2019anomaly,morais2019learning,sultani2018real}, etc. While these problems are being widely studied, a related problem, detecting deception, has not been the focus of much research.

Masip et al.~\cite{MasipDefiningDeception} define deception as ``the deliberate attempt, whether successful or not, to conceal, fabricate, and/or manipulate in any other way, factual and/or emotional information, by verbal and/or nonverbal means, in order to create or maintain in another or others a belief that the communicator himself or herself considers false''. Motives for deceptive behaviors can vary from inconsequential to those constituting a serious security threat. Many applications related to computer vision, human-computer interfaces, security, and computational social sciences need to be able to automatically detect such behaviors in public areas (airports, train stations, shopping malls), simulated environments, and social media~\cite{tsikerdekis2014online}. In this paper, we address the problem of automatically detecting deceptive behavior learned from gaits and gestures from walking videos.


Deception detection is a challenging task because deception is a subtle human behavioral trait, and deceivers attempt to conceal their actual cues and expressions. However, there is considerable research on verbal (explicit) and nonverbal (implicit) cues of deception~\cite{EkmanNonverbalLeakage}. Implicit cues such as facial and body expressions~\cite{EkmanNonverbalLeakage}, eye contact~\cite{Zuckerman1981}, and hand movements~\cite{JensenIdentification} can provide indicators of deception. Facial expressions have been widely studied as cues for automatic recognition of deception~\cite{ding2019face,michael2010motion,meservy2005deception,perez2015verbal,perez2015deception,hu2018deep,wu2018deception}. However, deceivers try to alter or control what they think is getting the most attention from others~\cite{EkmanTellingLies1985}. Compared to facial expressions, body movements such as gaits are more implicit and are less likely to be controlled. This makes gait an excellent avenue for observing deceptive behaviors. The psychological literature on deception also shows that multiple factors need to be considered when deciphering nonverbal information~\cite{Buller1996}.

\textbf{Main Results:} We present an data-driven approach for detecting deceptive walking of individuals based on their gaits and gestures as extracted from their walking videos (Figure~\ref{fig:cover}). Our approach is based on the assumption that humans are less likely to alter or control their gaits and gestures than facial expressions~\cite{EkmanTellingLies1985}, which arguably makes such cues better indicators of deception. 


Given a video of an individual walking, we extract his/her gait as a series of 3D poses using state-of-the-art human pose estimation~\cite{dabral2018learning}.  We also annotate various gestures performed during the video. Using this data, we compute psychology-based gait features, gesture features, and deep features learned using an LSTM-based neural network. These gait, gesture, and deep features are collectively referred to as the \textit{deceptive features}. Then, we feed the deceptive features into fully connected layers of the neural network to classify normal and deceptive behaviors. We train this neural network classifier (\textit{Deception Classifier}) to learn the deep features as well as classify the data into behavior labels on a novel dataset (\textit{DeceptiveWalk}). Our Deception Classifier can achieve an accuracy of \accuracy~when classifying deceptive walking, which is an improvement of $4.7\%$ and $9.2\%$ over classifiers based on the state-of-the-art emotion~\cite{bhattacharya2019step} and action~\cite{shi2019skeleton} classification algorithms, respectively.

Additionally, we present our deception dataset, \textit{DeceptiveWalk}, which contains $2224$ annotated gaits and gestures collected from $162$ individuals performing deceptive and natural walking. The videos in this dataset provide interesting observations about participants' behavior in deceptive and natural conditions. Deceivers put their hands in their pockets and look around more than the participants in the natural condition. Our observations also corroborate previous findings that deceivers display the opposite of the expected behavior or display a controlled movement~\cite{EkmanNonverbalLeakage,DePauloCues,Navarro2003}. 


Some of the novel components of our work include:\\
\noindent 1. A novel deception feature formulation of gaits and gestures obtained from walking videos based on psychological characterization and deep features learned from an LSTM-based neural network.

\noindent 2. A novel deception detection algorithm that detects deceptive walking with an accuracy of \accuracy.

\noindent 3. A new public domain dataset, \textit{DeceptiveWalk}, containing annotated gaits and gestures with deceptive and natural walks.

The rest of the paper is organized as follows. In Section 2, we give a brief background on deception and discuss previous methods of detecting deceptive walking automatically. In Section 3, we give an overview of our approach and present the details of our user study. We also present our novel \textit{DeceptiveWalk} dataset in Section 3. We present the novel deceptive features in Section 4 and provide details of our  method for detecting deception from walking videos. We highlight the results generated using our method in Section 5.

%% file: 2_related.tex
\section{Related Work}
In this section, we give a brief background of research on deception and discuss previous methods for detecting deceptive behaviors automatically.

\subsection{Deception}
Research on deception shows that people behave differently when they are being deceitful, and different clues can be used to detect deception~\cite{EkmanNonverbalLeakage}. Darwin first suggested that certain movements are ``expressive'' and escape the controls of the will~\cite{Darwin}. Additionally, when people know that they are being watched, they alter their behavior in something known as ``The Hawthorne Effect''~\cite{LevittHawthorneEffect}. To avoid detection, deceivers try to alter or control what they think others are paying the most attention to~\cite{EkmanTellingLies1985}. Different areas of the body have different communicating abilities that provide information based on differences in movement, visibility, and speed of transmission~\cite{EkmanNonverbalLeakage}. Therefore, using channels to which people pay less attention and that are harder to control are good indicators of deception. 

Body expressions present an alternative channel for perception and communication, as shown in emotion research~\cite{Aviezer}. Body movements are more implicit and may be less likely to be controlled compared to facial expressions, as evidenced by prior work, suggesting that clues such as less eye contact~\cite{Zuckerman1981}, downward gazing (through the affective experiences of guilt/sadness)~\cite{BaumeisterGuilt,EkmanTellingLies1985}, and general hand movements~\cite{JensenIdentification} are good indicators of deception. Though there is a large amount of research on non-verbal cues of deception, there is no single conclusive universal cue~\cite{vrij2019reading}. To address this issue, we present a novel data-driven approach that computes features based on gaits and gestures, along with deep learning.




\subsection{Automatic Detection of Deception}
Many approaches have been developed to detect deception automatically using verbal or text-based cues. Text-based approaches have been developed to detect deception in online communication~\cite{zhou2008following}, news~\cite{conroy2015automatic}, court cases~\cite{fornaciari2013automatic}, etc. A large number of approaches that detect deception using non-verbal cues have focused on facial expressions and head and hand gestures~\cite{perez2015verbal,perez2015deception,ding2019face,hu2018deep}. Some approaches use only facial expressions to detect deception~\cite{avola2019automatic,hasan2019facial}, while other approaches use only hand and head gestures~\cite{michael2010motion,meservy2005deception}. In situations where the subject is sitting and talking, Van der Zee et al.~\cite{van2019freeze} used full-body features captured using motion capture suits to detect deception. However, such an approach is impractical for most applications. Other cues that have been used for detecting deception include eye movements~\cite{zuo2019your}, fMRI~\cite{cui2013detection}, thermal input~\cite{buddharaju2005automatic,abouelenien2014deception}, and weight distribution~\cite{atlas2005detection}. Recent research has also introduced multimodal approaches that use multiple modalities such as videos, speech, text, and physiological signals to detect deception~\cite{gupta2019bag,rill2019high,wu2018deception,krishnamurthy2018deep,karimi2018toward,abouelenien2017gender,abouelenien2016detecting}. However, most of these approaches have focused on humans that have been sitting or standing; deception detection from walking is relatively unexplored. This is an important problem for many applications such as security in public areas (e.g., airports, concerts, etc.) and surveillance. Therefore, in this paper, we present an approach that detects deception from walking using gait and gesture features.

\subsection{Non-verbal Cues of Gaits and Gestures}
Research has shown that body expressions, including gaits and gestures, are critical for the expression and perception of others' emotions, moods, and intentions~\cite{kleinsmith2013affective}. Gaits have been shown to be useful in conveying and recognizing identity~\cite{wan2019survey}, gender~\cite{yu2009study}, emotions~\cite{kleinsmith2013affective}, moods~\cite{michalak2009embodiment}, and personalities~\cite{atkinson2007evidence,roether2009critical}. Gestures have also been widely observed to convey emotions~\cite{labarre1947cultural,noroozi2018survey}, intentions~\cite{mount2008intentions}, moods~\cite{de2006towards}, personality~\cite{ball1999relating}, and deception~\cite{michael2010motion,meservy2005deception,ding2019face,hu2018deep}. For gait recognition, many deep learning approaches have been proposed~\cite{zhang2019gait,zhang2019learning,wang2019ev}, and LSTM-based approaches have been used to model gait features~\cite{zhang2019gait}. Body expressions have also been used for anomaly detection in videos with multiple humans~\cite{morais2019learning}. Many deep learning approaches have been proposed to recognize and generate actions from 2D~\cite{khodabandeh2018diy,yang2018pose} and 3D skeleton data~\cite{shi2019skeleton,pavllo2018quaternet,habibie2017recurrent}. Graph convolution networks such as STEP~\cite{bhattacharya2019step} and ST-GCN~\cite{yan2018spatial} for emotion and action recognition from skeleton data respectively have also been proposed. Inspired by these approaches that showcase the communicative capability of gaits and gestures, we use these non-verbal cues to detect deceptive walks.





%% file: 3_approach.tex
\section{Approach}
In this section, we give an overview of our approach. We present details of our user study that was used to derive a data-driven metric.

\subsection{Overview}
\begin{figure}[t]
      \centering
      \includegraphics[height=3.5cm]{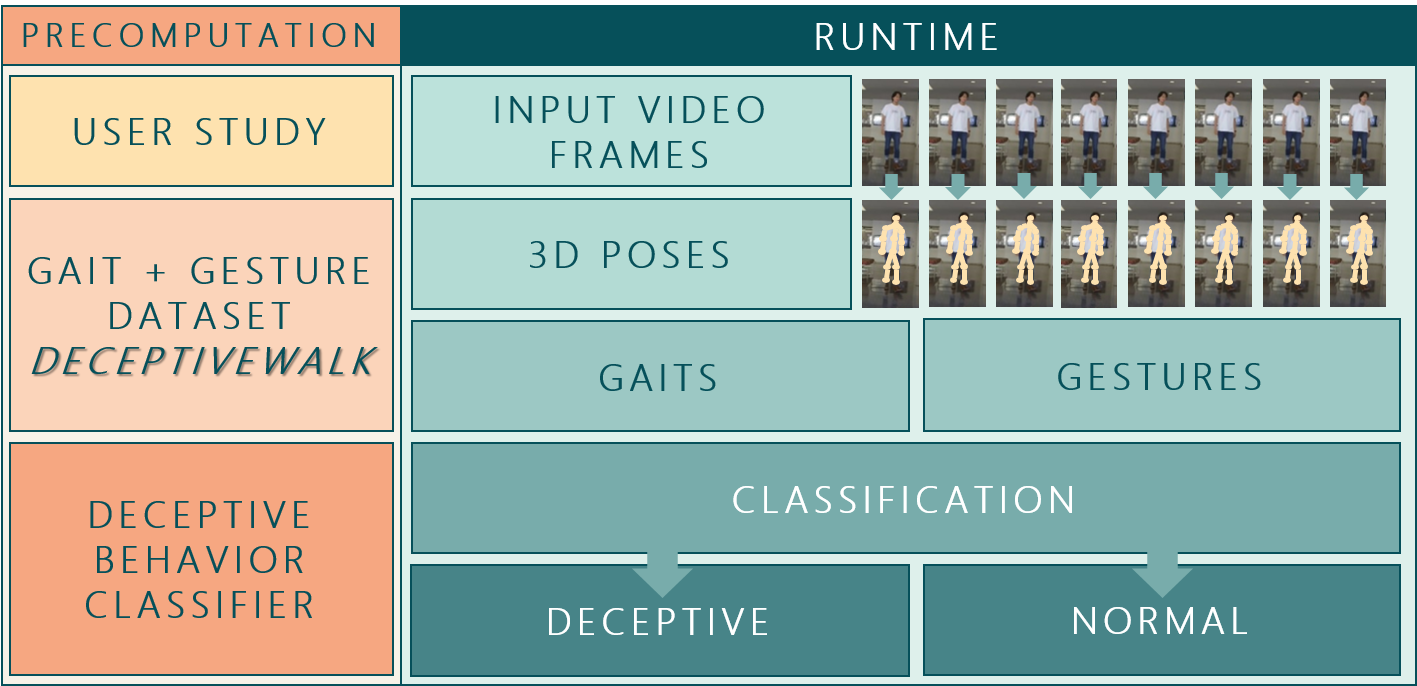}
      \caption{\textbf{Overview}: At runtime, we compute gait and gesture features from an input walking video using 3D poses. Using our novel DeceptiveWalk dataset, we train a classifier consisting of an LSTM module that is used to learn temporal patterns in the walking styles, followed by a fully connected module that is used to classify the features into class labels (natural or deceptive).}
      \vspace{-15pt}
      \label{fig:overview}
\end{figure}

We provide an overview of our approach in Figure~\ref{fig:overview}. Our data-driven algorithm consists of an offline training phase during which we conducted a user study and obtained a video dataset of participants performing either deceptive or natural walks. For each video, we use a state-of-the-art 3D human pose extraction algorithm to extract gaits as a series of 3D poses. Using an extracted pose decouples the problem of interpreting the poses from the problem of inferring the behavior characteristics of the individuals in the videos. Similar choices have also been made for action recognition (DGNN~\cite{shi2019skeleton}) and sentiment analysis (STEP~\cite{bhattacharya2019step}). We also annotate various hand and head gestures performed by participants during the video. We compute psychology-based gait features, gesture features, and deep features learned using an LSTM-based neural network. Using these features with their deception labels, we train a \textit{Deception Classifier} that detects deceptive walks. At runtime, our algorithm takes a video as input and extracts gaits and gestures. Using the trained Deception Classifier, we can then detect whether the individual in the video is performing a deceptive walk or not. We now describe each component of our algorithm in detail.

\subsection{Notation}
We represent the deception dataset by $\mathbb{D}$. We obtain this DeceptiveWalk dataset from a data collection study. We denote a data point in this dataset by $M_i$, where $i \in \{1, 2, ..., N\}$ and $N$ is the number of gaits in the dataset. A data point $M_i$ contains the gait $W_i$, the gestures $G_i$ extracted from the $i^{th}$ video, and its associated deception label $d_i \in \{0, 1\}$. A value of $d_i = 0$ represents a natural walking video and $d_i = 1$ represents a deceptive walking video.

We use a joint representation of humans for our gait formulation. Similar to the previous literature on modeling human poses~\cite{dabral2018learning}, we use a set of $16$ joints to represent the poses. A set of 3D positions of each joint $j_i, i \in \{1,2, ..., 16\}$ represents a human pose. We define the gait $W_i$ of a human obtained from the $i^{th}$ video as a series of poses $P_i^{k} \in \mathbb{R}^{48}$ where $k \in [1, 2, ..., \tau]$. Each pose corresponds to a frame $k$ from the video, and $\tau$ is the total number of frames in the video.

\subsection{Data Collection Study}
We conducted a user study to collect gait and gesture data of individuals performing either deceptive or natural walks. We designed this study after working closely with social psychologists. Many previous studies have demonstrated that participants respond in a naturalistic manner in experimentally controlled settings (e.g., classic studies by Milgram~\cite{milgram1978obedience} and Asch~\cite{asch1956studies}). Motivated by these approaches, our approach also induces naturalistic deceptive behaviors induced via the experimenter's instructions during the briefing.

\subsubsection{Participants}
We recruited $162$ participants ($109$ female, $49$ male, $4$ preferred not to say, $\overline{age} = 20.39$) from a university campus. Appropriate institutional review board (IRB) approval was obtained for the study and each participant provided informed written consent to participate in the study and record videos.

\subsubsection{Procedure}
\begin{figure}[t]
      \centering
      \includegraphics[width=\linewidth]{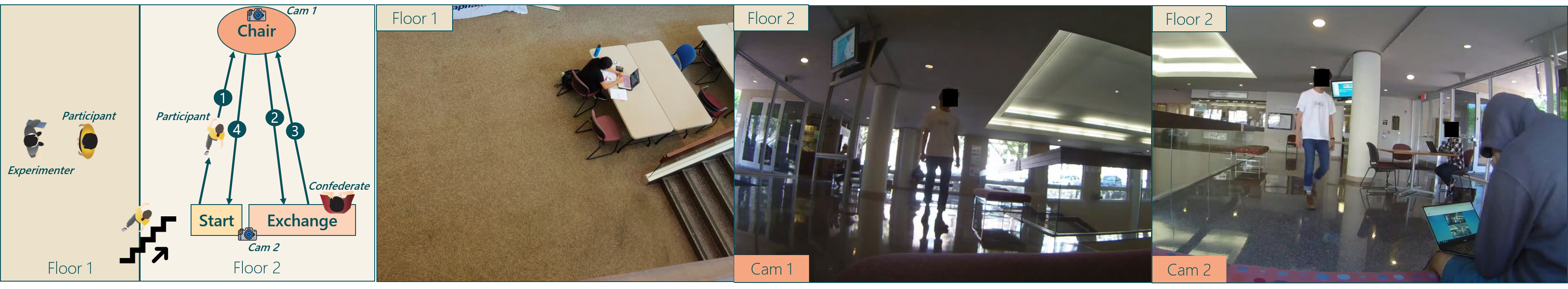}
      \caption{\textbf{Experiment Area}: The experiment area consists of two floors. The experimenter briefs the participant on Floor 1, and the participant performs the task on Floor 2. We obtain four videos of walking captured by two cameras, Cam 1 and Cam 2, for each participant. We show screenshots of a participant performing the task from Cam 1 and Cam 2.}\label{fig:exp}
      \vspace{-15pt}
\end{figure}

We adopted a between-subject method for data collection. We compared two conditions: \textit{natural} and \textit{deceptive}. Each participant was randomly assigned to walk either naturally or in a deceptive manner, as described below. In both conditions, the experiment area remains the same. We show the experiment setup in Figure~\ref{fig:exp}. The experiment consists of the following steps:

\begin{enumerate}
    \item The experiment area consists of two floors. The participant is briefed about the procedure by the experimenter on Floor 1. The participant is informed that he/she is supposed to walk to a chair, pick up an object from the chair, exchange it with a person wearing a hoodie, replace the package on the chair, and come back.
    \item The participant then takes the stairs from Floor 1 to Floor 2 and emerges at the \textit{Start} location. 
    \item The participant then walks towards the \textit{Chair}. We refer to this walk as Walk 1, and it is captured by the camera placed behind the chair (Cam 1).
    \item After reaching the \textit{Chair}, the participant picks up an object. In the natural condition, the object is a collection of classic computer science papers and is placed on top of the chair. However, in the deceptive condition, the object is a big roll of money and is placed under the chair.
    \item The participant then walks towards the \textit{Exchange} location. We refer to this walk as Walk 2, and it is captured by the camera placed behind the \textit{Exchange/Start} location (Cam 2).
    \item At the \textit{Exchange} location, the participant exchanges the object with a package from the confederate sitting there. The package is a red package that remains the same in both conditions. In the briefing, the participant is informed that it is a piece of computer hardware.
    \item After the exchange, the participant walks back towards the \textit{Chair}. We refer to this walk as Walk 3, and it is captured by the camera placed behind the chair (Cam 1).
    \item The participant places the exchanged package on the chair.
    \item The participant walks towards the stairs located near the \textit{Start} location. We refer to this walk as Walk 4, and it is captured by the camera placed behind the chair (Cam 2).
    \item The participant walks towards the experimenter on Floor 1 and is debriefed about the experiment.
\end{enumerate}

In this procedure, the deceptive walk is induced by the experimenter during the briefing, the type of object, and the appearance of the confederate. The experiment was conducted in a public setting (lobby of an academic building in a university). Throughout the experiment, there were $4-10$ students sitting around the lobby area who were not part of the experiment. In the deceptive condition, the participant is supposed to conceal their activities from these ``onlookers.'' The experimenter stresses that the package is a secret, and because of the amount of money involved, it is really important that the participant keeps anyone from noticing as he/she does these things. In the natural condition, the confederate is wearing a hoodie with the hood pulled down. However, in the deceptive condition, the confederate is wearing the same hoodie with the hood pulled up while wearing sunglasses. Because of this setting, the participant makes a deliberate attempt to conceal his/her activities, resulting in deceptive behavior~\cite{MasipDefiningDeception}. We present sample videos containing deceptive and natural walks in the accompanying video.



\subsubsection{Data Labeling}
As described in the procedure, we obtain four walking videos for each participant. For each video, we extract the 3D pose of the participant in each frame using the human pose extraction algorithm described below. Specifically, we obtain the 3D position of each joint relative to the root joint in each frame. Depending on the condition, natural or deceptive, we assign a label to each walking video.

\subsection{Human Pose Extraction}\label{sec:poseExtraction}
To extract a person's poses from their walking video, we first need to distill out extraneous information such as attire, items carried (\textit{e.g.}, bags or cases), background clutter, etc. We adapt the approach of~\cite{dabral2018learning}, where the authors have trained a weakly supervised network to perform this task. The first part of the network, called the \textit{Structure-Aware PoseNet (SAP-Net)}, is trained on spatial information, which learns to extract 3D joint locations from each frame of an input video. The second part of the network, called the \textit{Temporal PoseNet (TP-Net)}, is trained on temporal information, which takes in the extracted joints and outputs a temporally harmonized sequence of poses.

Moreover, walking videos are collected from various viewpoints, and the scale of the person varies depending on the relative camera position. Therefore, we perform a least-squares similarity transform~\cite{umeyama} on each pose in the dataset. This step ensures that each individual pose lies within a box of unit volume, and the first pose of each video is centered at the global origin.

\subsection{Gestures}\label{sec:gestures}
In addition to extracting gaits, we also annotate the gestures performed by the participants during the four walks. Prior literature on deception suggests that people showing deceptive behavior often feel distressed, and levels of discomfort can be used to detect a person's truthfulness. These levels of discomfort may appear in fidgeting (adjusting their shirt/moving their hands) or while glancing at objects such as a clock or a watch~\cite{Navarro2003}. Touching the face around the forehead, neck, or back of the head is also an indicator of discomfort related to deception~\cite{Givens2002}. We use these findings and consider the following set of gestures:\{Hands In Pockets, Looking Around, Touching Face, Touching Shirt/Jackets, Touching Hair, Hands Folded, Looking at Phone\}. We chose this set because it includes all the hand gestures observed in the walking videos of participants, and these gestures have been reported to be related to deception~\cite{Givens2002,Navarro2003}. For each walking video, we annotate whether each gesture from this set is present or absent. For the \textit{hands in pockets} gesture, we also annotate how many hands are in the pocket.


\subsection{DeceptiveWalk Dataset}
We invited $162$ participants for the data collection study. For each participant, we obtained four walking videos. Some participants followed the instructions incorrectly, and we could not obtain all four walking videos for these participants. Overall, we obtained $589$ walking videos. For each video, we extracted gaits using the human pose extraction algorithm. Due to occlusions, the human pose extraction algorithm had significant errors in $33$ of these videos. To expand the size and diversity of the dataset, we performed data augmentation by reflecting all the 3D pose sequences about the vertical axis and performing phase shifts in the temporal domain. Reflection about the vertical axis and the phase shift does not alter the overall gaits. Hence the corresponding labels can remain the same. As a result, we were able to obtain a total of $2224$ gaits of participants from the data collection study. 

Depending on the condition assigned to a participant, we assigned each video with a label of natural or deceptive. We also annotated the various gestures from the gesture set performed by the participants in each video. We refer to this dataset of $2224$ gaits, gestures, and their associated deception labels as the DeceptiveWalk dataset. The dataset contains $1004$ natural and $1220$ deceptive videos. The dataset contains $552$ videos of Walk 1, $564$ videos of Walk 2, $532$ videos of Walk 3, and $576$ videos of Walk 4. We will make this dataset publicly available for future research.

%% file: 4_classification.tex
\section{Automatic Deception Detection}
From the user study, we obtain a dataset of 3D pose data for each video. Using this pose data, we extract gait and gesture features. We also compute deep features using a deep LSTM-based neural network. We use these novel deception features as input to a classification algorithm. In this section, we first describe these deception features and their computation. We then describe the classification algorithm.

\subsection{Gait Features}
\begin{table}[t]
\caption{\textbf{Posture Features}: In each frame, we compute the following features that represent the human posture.}
\centering
\begin{tabular}{|l|l|}
\hline
\multicolumn{1}{|c|}{Type} & \multicolumn{1}{c|}{Description} \\ \hline
Volume & Bounding box \\ \cline{1-2}
\multirow{5}{*}{Angle At}
                                & Neck by shoulders \\ \cline{2-2}
                                & Right shoulder by neck and left shoulder \\ \cline{2-2}
                                & Left shoulder by neck and right shoulder\\ \cline{2-2}
                                & Neck by vertical and back  \\ \cline{2-2}
                                & Neck by head and back  \\ \cline{1-2}
\multirow{5}{*}{\begin{tabular}[c]{@{}l@{}}Distance \\ Between\end{tabular}}                                                           & Right hand and the root joint \\ \cline{2-2}
                                & Left hand and the root joint \\ \cline{2-2}
                                & Right foot and the root joint \\ \cline{2-2}
                                & Left foot and the root joint \\ \cline{2-2}
                                & Consecutive foot strikes (stride length) \\ \cline{1-2}
\multirow{2}{*}{\begin{tabular}[c]{@{}l@{}}Area of \\ Triangle\end{tabular}}                                                           & Between hands and neck \\ \cline{2-2}
                                & Between feet and the root joint \\ \hline
\end{tabular}
\vspace{-15pt}
\label{tab:posturefeatures}
\end{table}

In previous work~\cite{randhavane2019identifying}, researchers have used a combination of posture and movement features to represent a gait. This is based on the finding that both posture and movement features are used to predict an individual's affective state~\cite{kleinsmith2013affective}. These features relate to the joint angles, distances between the joints, joint velocities, and space occupied by various parts of the body and have been used to classify gaits according to emotions and affective states ~\cite{crenn2016body,kleinsmith2013affective,randhavane2019identifying}. Since deceptive behavior impacts the emotional and cognitive state of an individual~\cite{Slepian,EkmanNonverbalLeakage}, features that are indicators of affective state may also be related to deception. Therefore, we use a similar set of gait features for the classification of deceptive walks, as described below. Combining the posture and the movement features, we obtain $29$ dimensional gait features.



\subsubsection{Posture Features}
In each frame, we compute the features relating to the distances between joints, angles between joints, and the space occupied by various parts of the body. These features correspond to the body posture in that frame. We use the 3D joint positions computed using the human pose extraction algorithm (Section~\ref{sec:poseExtraction}) to compute these posture features as described below:
\begin{itemize}[noitemsep,topsep=0pt]
    \item Volume: We use the volume of the bounding box around a human as the feature that represents the compactness of the human's posture.
    \item Area: In addition to volume, we also use the areas of triangles between the hands and the neck and between the feet and the root joint to model body compactness.
    \item Distance: We use the distances between the feet, the hands, and the root joint as features. These features model body expansion and also capture the magnitude of the hand and food movement.
    \item Angle: We use the angles extended by different joints at the neck to capture the head tilt and rotation. These features also capture whether the posture is slouched or erect using the angle extended by the shoulders at the neck.
    \item Stride Length: Stride length has been used to represent gait features in literature; therefore, we also include stride length as a posture feature. We compute the stride length by computing the maximum distance between the feet across the gait.
\end{itemize}
We summarize these posture features in Table~\ref{tab:posturefeatures}. There are $13$ posture features in total.

\begin{figure}[t]
    \centering
    \includegraphics[width=\columnwidth]{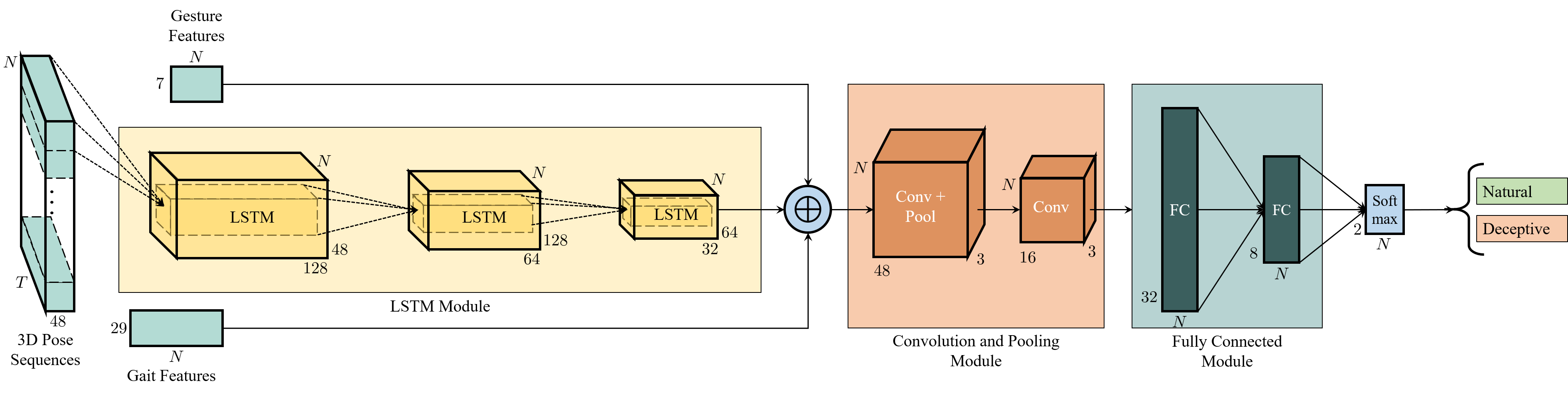}
    \caption{\textbf{Our Classification Network}: Each of the 3D pose sequences corresponding to various walking styles is fed as input to the LSTM module consisting of $3$ LSTM units, each of depth $2$. The output of the LSTM module (the deep feature pertaining to the 3D pose sequence) is concatenated with the corresponding gait feature and the gesture feature, denoted by the $\mathbf{\bigoplus}$ symbol. The concatenated features are then fed into the fully connected module, which consists of two fully connected layers. The output of the fully connected module is passed through another fully connected layer equipped with the softmax activation function to generate the predicted class labels.}
    \vspace{-15pt}
    \label{fig:neural_net}
\end{figure}

\subsubsection{Movement Features}

In addition to the human posture in each frame, movement of the joints across time is also an important feature of gaits~\cite{kleinsmith2013affective}. We model this by computing the magnitude of the speed, acceleration, and movement jerk of the hands, head, and foot joints.  For each joint, we compute these features using the first, second, and third finite derivatives of its 3D position computed using the human pose extraction algorithm. We also include the gait cycle time as a feature. We compute this by the time between two consecutive foot strikes of the same foot. There are $16$ movement features in total. We aggregate both posture and movement features over the gait to form the $29$ dimensional gait feature vector.



\subsection{Gesture Features}
We use the set of gestures described in Section~\ref{sec:gestures} and formulate the gesture features as a $7$ dimensional vector corresponding to the set \{Hands In Pockets, Looking Around, Touching Face, Touching Shirt/Jackets, Touching Hair, Hands Folded, Looking at Phone\} $\in \mathbb{R}^7$. For each gesture $j$ in the set, we set the value of $G_i^j = 1$ if the gesture is present in the walking video and $G_i^j = 0$  if it is absent. For \textit{hands in pockets}, we use $G_i^j = 1$ if one hand is in the pocket, $G_i^j = 2$ if two hands are in the pocket, and $G_i^j = 0$  if no hands are in the pockets. 

\subsection{Deep Features}
Given a sequence of 3D pose data for a fixed number of time frames as input, the task of the classification algorithm is to assign the input one of two class labels --- natural or deceptive. To achieve this, we develop a deep neural network consisting of \textit{long short-term memory} (LSTM)~\cite{lstm} layers. LSTM units consist of feedback connections and gate functions that help them retain useful patterns from input data sequences. Since the inputs in our case are walking motions, which are periodic in time, we reason that LSTM units can learn to efficiently encode the different walking patterns in our data, which, in turn, helps the network segregate the data into the respective classes. We call the feature vectors learned from the LSTM layers, \textit{deep features}, $f_d \in \mathbb{R}^{32}$.

\subsection{Classification Algorithm and Network Architecture}
Our overall neural network is shown in Figure~\ref{fig:neural_net}. We first normalize the input sequences of 3D poses so that each individual value lies within the range of $0$ and $1$. We feed the normalized sequences of 3D poses into an LSTM module, which consists of $3$ LSTM units of sizes $128$, $64$, and $32$, respectively, and each of depth $2$. We concatenate the $32$ dimensional deep features (output from the LSTM module) with the $29$ dimensional gait and the $7$ dimensional gesture features and feed the $68$ dimensional combined deceptive feature vectors into a convolution and pooling module. This module consists of $2$ convolution layers. The first convolution layer has a depth of $48$ and a kernel size of $3$. It is followed by a maxpool layer with a window size of $3$. The second convolution layer has depth $16$ and a kernel size of $3$. The output of the second convolution layer is flattened and passed into the fully connected module, which consists of $2$ fully connected (FC) layers of sizes $32$ and $8$, respectively. All the FC layers are equipped with the ELU activation function. The output feature vectors from the fully connected module are passed through a $2$ dimensional fully connected layer with the softmax activation function to generate the output class probabilities. We assign the predicted class label as the one with a higher probability.

%% file: 5_results.tex

\section{Results}
We first describe the implementation details of our classification network, followed by a detailed summary of the experimental results.

\begin{table}[t]
\caption{\textbf{Accuracy and Ablation Study}: We compared our method with state-of-the-art methods for gait-based action recognition as well as perceived emotion recognition. Both these classes of methods use similar gait datasets as inputs but learn to predict different labels. Additionally, we evaluated the usefulness of the different features for classifying a walk as natural or deceptive. We observed that all three components of the deceptive features (gait features, gesture features, and deep features) contribute towards the accurate prediction of deceptive behavior.}
\label{tab:accuracy}
\begin{tabular}{|c|c|c|c|c|c|c|c|}
\hline
\multirow{2}{*}{ST-GCN~\cite{yan2018spatial}} & \multirow{2}{*}{DGNN~\cite{shi2019skeleton}} & \multirow{2}{*}{STEP~\cite{bhattacharya2019step}} & \multicolumn{5}{c|}{Ours}                                \\ \cline{4-8} 
                        &                       &                       & Gestures & Gait    & Gestures + Gait & Deep    & All     \\ \hline
77.82\%                 & 79.19\%               & 83.68\%               & 61.59\%  & 72.56\% & 77.74\%         & 82.67\% & \accuracybold \\ \hline
\end{tabular}
\end{table}


\subsection{Implementation Details}
We randomly selected $80\%$ of the dataset for training the network, $10\%$ for cross-validation and kept the remaining $10\%$ for testing. Inputs were fed to the network with a batch size of $8$. We used the standard cross-entropy loss to train the network. The loss was optimized by running the Adam optimizer~\cite{adam} for $n = 500$ epochs with a momentum of $0.9$ and a weight decay of $10^{-4}$. The initial learning rate was $0.001$, which was reduced to half its present value after $250$ $\left(\frac{n}{2}\right)$, $375$ $\left(\frac{3n}{4}\right)$, and $437$ $\left(\frac{7n}{8}\right)$ epochs.

\subsection{Experiments}
We evaluate the performance of our LSTM-based network as well as the usefulness of the deep features through exhaustive experiments. All the experimental results are summarized in Table~\ref{tab:accuracy}.


We compare the performance of using the LSTM-based network with the various input features individually. Gesture features by themselves provide the lowest classification accuracy because they only coarsely summarize the subject's activities and not their walking patterns. Gait features, on the other hand, contain only this information and are seen to be more helpful in distinguishing between the class labels. Gestures and gait features collectively perform better than their individual performances. However, these features still lose some of the useful temporal patterns in the original 3D pose sequences. Using LSTMs to learn the temporal patterns directly from the 3D pose sequences, we observe an improvement of $\sim 5\%$ over using the combined gait and the gesture features. Finally, combining the deep features learned from the LSTM module with the gait and gesture features leads to an overall classification accuracy of \accuracy.

Since ours is the first algorithm that detects deception from walking gaits, we compared our method with prior methods for emotion and action recognition from gaits since these methods also solve the problem of gait classification (albeit with a different set of labels). We compare with approaches that use spatial-temporal graph convolution networks for emotion (STEP~\cite{bhattacharya2019step}) and action recognition (ST-GCN~\cite{yan2018spatial}). We also compare our method with an approach that uses a novel directed graph neural network (DGNN~\cite{shi2019skeleton}) for action recognition. We train these methods on our DeceptiveWalk dataset and obtain their performance on the testing set similar to our method. Our method outperforms these state-of-the-art models used for emotion and action recognition by a minimum of $4.7\%$ (Table~\ref{tab:accuracy}). We also conducted a subject independent study where the train and test datasets had different sets of participants. We observed accuracy of $85.06\%$ indicating that the accuracy of our algorithm does not depend on the participants.

\begin{figure}[t]
    \centering
    \includegraphics[width=0.8\columnwidth]{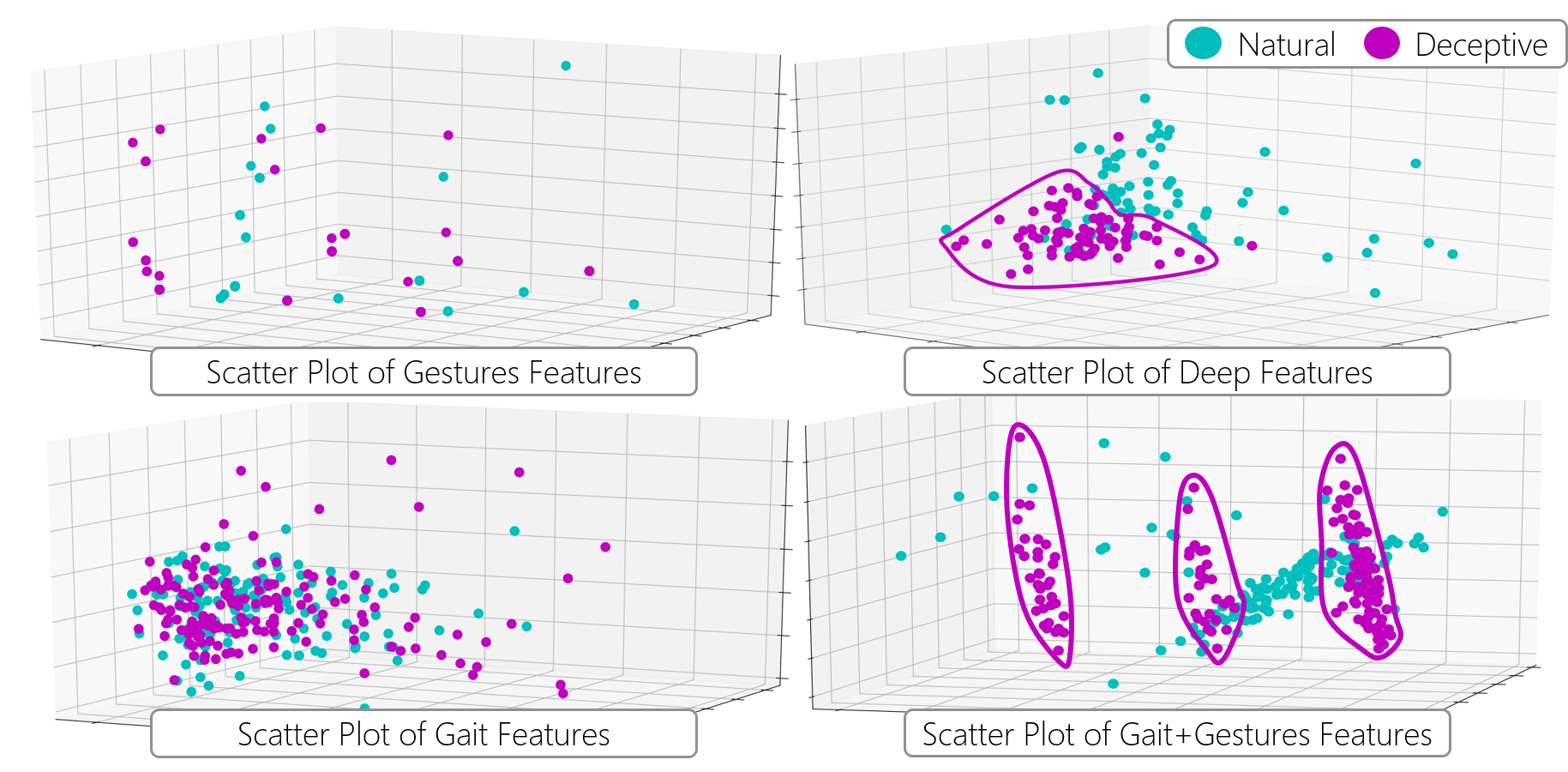}
    \caption{\textbf{Separation of Features}: We present the scatter plot of the features for each of the two-class labels. We project the features to a $3$ dimensional space using PCA for visualization. The gait and gesture features are not separable in this space. However, the combined gait+gestures features and deep features are well separated even in this low dimensional space. The boundary demarcated for the deceptive class features is only for visualization and does not represent the true class boundary.}
    \label{fig:scatter}
\end{figure}

Furthermore, we show the scatter of the features from both the natural and the deceptive classes in Figure~\ref{fig:scatter}. The original features are high dimensional ($f_{gait} \in \mathbb{R}^{29}, f_{gesture} \in \mathbb{R}^{7}$, and $f_d \in \mathbb{R}^{32}$). Hence we perform PCA to project and visualize them on a $3$ dimensional space. The gait and gesture features from the two classes are not separable in this space. However, the combined gait+gesture features, as well as the deep features from the two classes, are well-separated even in this lower-dimensional space, implying that the deep features ($f_d$) and gait+gesture features ($f_g$) can efficiently separate between the two classes.

\subsection{Analysis of Gesture Cues}

\begin{figure}[t]
    \centering
    \includegraphics[width=0.75\columnwidth]{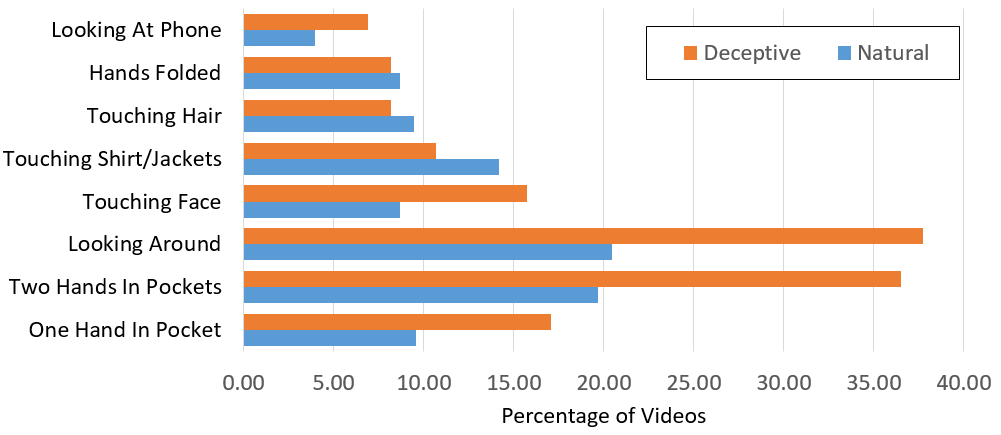}
    \caption{\textbf{Gesture Features}: We present the percentage of videos in which the gestures were observed for both deceptive and natural walks. We observe that deceivers put their hands in their pockets and look around more than participants in the natural condition. However, the trend is unclear in other gestures. This could be explained in previous literature that suggests that deceivers try to alter or control what they think others are paying the most attention to~\protect\cite{EkmanNonverbalLeakage,DePauloCues,Navarro2003}.}
    \label{fig:gestureDist}
    \vspace{-15pt}
\end{figure}

We tabulate the distribution of various gestures in Figure~\ref{fig:gestureDist}. We can make interesting observations from this data. 
\begin{enumerate}
    \item Deceivers are more likely to put their hands in their pockets, look around and check if anyone is noticing them, touch their face, and look at their phone than the participants in the natural condition.
    \item Participants in the natural condition touched their shirt/jacket more than the deceivers.
    \item Deceivers touched their hair and folded their hands at about the same rate as those in the natural condition.
\end{enumerate}
Previous literature suggests that deceivers are sometimes very aware of the cues they are putting out and may, in turn, display the opposite of the expectation or display a controlled movement~\cite{EkmanNonverbalLeakage,DePauloCues,Navarro2003}. This would explain observations 2 and 3. Factors like how often a person lies or how aware they are of others' perceptions of them could contribute to whether they show fidgeting and nervous behavior or controlled and stable movements. These results are in accordance with the Hawthorne effect. However, more analysis is necessary to accurately predict the importance of each gesture in the prediction and expression of deception in real-world cases.

%% file: 6_conclusion.tex
\section{Conclusion, Limitations, and Future Work}
We presented a novel method for distinguishing the deceptive walks of individuals from natural walks based on their walking style or pattern in videos. Based on our novel \textit{DeceptiveWalk} dataset, we train an LSTM-based classifier that classifies the \textit{deceptive features} and predicts whether an individual is performing a deceptive walk. We observe an accuracy of \accuracy~obtained using 10-fold cross-validation on $2224$ data points.


There are some limitations to our approach. Our approach is based on our dataset collected in a controlled university lab setting. Participants in the study were aware of the context of the experiment, and compared to real-world situations, the stakes were smaller in our experiment. This means that our results are preliminary and may not be directly applicable for detecting many kinds of deceptive walking behaviors in the real-world. We need considerable more investigation and research in terms of testing these ideas in all kind of situations. This would require collecting a large sample of unbiased video data in different real-world scenarios and analyzing them. While we present an algorithm that detects deception using walking, we do not claim that performing certain gestures and walking in a certain way conveys deception in all cases. Additionally, since our algorithm does not provide $100\%$ accuracy for classifying both deceptive and natural walking, there can be false positives and false negatives. Therefore, other data (manual intervention and/or using other modalities) should be considered before assigning a final deceptive or a natural label to the walking when using our method in real-world applications.

Since the accuracy of our algorithm depends on the accurate extraction of 3D poses, the algorithm may perform poorly in cases where pose estimation is inaccurate (e.g., occlusions, bad lighting). For this work, we manually annotate the gesture data. In the future, we would like to automatically annotate different gestures performed by the individual and automate the entire method. Additionally, our approach is limited to walking. In the future, we would like to extend our algorithm to more general cases that include a variety of activities and also consider factors such as gender, culture, age, disabilities, etc. The analysis of our data collection study reveals an interesting mapping between gestures and deception. However, this mapping may depend on other factors and we would like to explore the factors that govern the relationship between gestures and deception. Finally, we would like to combine our method with other verbal and non-verbal cues of deception (e.g., gazing, facial expressions, etc.) and compare the usefulness of body expressions to other cues in detecting deception. 

%% file: main.bbl
\begin{thebibliography}{10}
\providecommand{\url}[1]{\texttt{#1}}
\providecommand{\urlprefix}{URL }
\providecommand{\doi}[1]{https://doi.org/#1}

\bibitem{abouelenien2014deception}
Abouelenien, M., P{\'e}rez-Rosas, V., Mihalcea, R., Burzo, M.: Deception
  detection using a multimodal approach. In: Proceedings of the 16th
  International Conference on Multimodal Interaction. pp. 58--65. ACM (2014)

\bibitem{abouelenien2016detecting}
Abouelenien, M., P{\'e}rez-Rosas, V., Mihalcea, R., Burzo, M.: Detecting
  deceptive behavior via integration of discriminative features from multiple
  modalities. IEEE Transactions on Information Forensics and Security
  \textbf{12}(5),  1042--1055 (2016)

\bibitem{abouelenien2017gender}
Abouelenien, M., P{\'e}rez-Rosas, V., Zhao, B., Mihalcea, R., Burzo, M.:
  Gender-based multimodal deception detection. In: Proceedings of the Symposium
  on Applied Computing. pp. 137--144. ACM (2017)

\bibitem{asch1956studies}
Asch, S.E.: Studies of independence and conformity: I. a minority of one
  against a unanimous majority. Psychological monographs: General and applied
  \textbf{70}(9), ~1 (1956)

\bibitem{atkinson2007evidence}
Atkinson, A.P., Tunstall, M.L., Dittrich, W.H.: Evidence for distinct
  contributions of form and motion information to the recognition of emotions
  from body gestures. Cognition  \textbf{104}(1),  59--72 (2007)

\bibitem{atlas2005detection}
Atlas, D., Miller, G.L.: Detection of signs of attempted deception and other
  emotional stresses by detecting changes in weight distribution of a standing
  or sitting person (Feb~8 2005), uS Patent 6,852,086

\bibitem{Aviezer}
Aviezer, H., Trope, Y., Todorov, A.: Body cues, not facial expressions,
  discriminate between intense positive and negative emotions. Science
  \textbf{228}(6111),  1225--1229 (2012)

\bibitem{avola2019automatic}
Avola, D., Cinque, L., Foresti, G.L., Pannone, D.: Automatic deception
  detection in rgb videos using facial action units. In: Proceedings of the
  13th International Conference on Distributed Smart Cameras. p.~5. ACM (2019)

\bibitem{ball1999relating}
Ball, G., Breese, J.: Relating personality and behavior: Posture and gestures.
  In: International Workshop on Affective Interactions. pp. 196--203. Springer
  (1999)

\bibitem{BaumeisterGuilt}
Baumeister, R.F., Stillwell, A.M., Heatherton, T.F.: Guilt: An interpersonal
  approach. Psychological Bulletin  \textbf{115}(2),  243--267 (1994)

\bibitem{bhattacharya2019step}
Bhattacharya, U., et~al.: Step: Spatial temporal graph convolutional networks
  for emotion perception from gaits. arXiv preprint arXiv:1910.12906  (2019)

\bibitem{buddharaju2005automatic}
Buddharaju, P., Dowdall, J., Tsiamyrtzis, P., Shastri, D., Pavlidis, I., Frank,
  M.: Automatic thermal monitoring system (athemos) for deception detection.
  In: 2005 IEEE Computer Society Conference on Computer Vision and Pattern
  Recognition (CVPR'05). vol.~2, pp. 1179--vol. IEEE (2005)

\bibitem{Buller1996}
Buller, D., Burgoon, J.: Interpersonal deception theory. Communication Theory
  \textbf{6}(3),  203--242 (1996)

\bibitem{conroy2015automatic}
Conroy, N.J., Rubin, V.L., Chen, Y.: Automatic deception detection: Methods for
  finding fake news. Proceedings of the Association for Information Science and
  Technology  \textbf{52}(1), ~1--4 (2015)

\bibitem{crenn2016body}
Crenn, A., Khan, A., et~al.: Body expression recognition from anim. 3d
  skeleton. In: IC3D (2016)

\bibitem{cui2013detection}
Cui, Q., Vanman, E.J., Wei, D., Yang, W., Jia, L., Zhang, Q.: Detection of
  deception based on fmri activation patterns underlying the production of a
  deceptive response and receiving feedback about the success of the deception
  after a mock murder crime. Social cognitive and affective neuroscience
  \textbf{9}(10),  1472--1480 (2013)

\bibitem{dabral2018learning}
Dabral, R., Mundhada, A., et~al.: Learning 3d human pose from structure and
  motion. In: ECCV (2018)

\bibitem{Darwin}
Darwin, C., Prodger, P.: The Expression of Emotions in Man and Animals.
  Philosophical Library (1972/1955)

\bibitem{de2006towards}
De~Silva, P.R., Osano, M., Marasinghe, A., Madurapperuma, A.P.: Towards
  recognizing emotion with affective dimensions through body gestures. In: 7th
  International Conference on Automatic Face and Gesture Recognition (FGR06).
  pp. 269--274. IEEE (2006)

\bibitem{DePauloCues}
DePaulo, B., Lindsay, J., Malone, B., Muhlenbruck, L., Charlton, K., Cooper,
  H.: Cues to deception. Psychological Bulletin  \textbf{129}(1),  74--118
  (2003)

\bibitem{ding2019face}
Ding, M., Zhao, A., Lu, Z., Xiang, T., Wen, J.R.: Face-focused cross-stream
  network for deception detection in videos. In: Proceedings of the IEEE
  Conference on Computer Vision and Pattern Recognition. pp. 7802--7811 (2019)

\bibitem{EkmanTellingLies1985}
Ekman, P.: Telling Lies: : Clues to deceit in the marketplace, marriage, and
  politics. Norton (1994)

\bibitem{EkmanNonverbalLeakage}
Ekman, P., Friesen, W.: Nonverbal leakage and clues to deception. Psychiatry
  \textbf{32}(1),  88--106 (1969)

\bibitem{fornaciari2013automatic}
Fornaciari, T., Poesio, M.: Automatic deception detection in italian court
  cases. Artificial intelligence and law  \textbf{21}(3),  303--340 (2013)

\bibitem{Givens2002}
Givens, D.: The Nonverbal Dictionary of Gestures, Signs, and Body Language
  Cues. Center for Nonverbal Studies Press (2002)

\bibitem{gupta2019bag}
Gupta, V., Agarwal, M., Arora, M., Chakraborty, T., Singh, R., Vatsa, M.:
  Bag-of-lies: A multimodal dataset for deception detection. In: Proceedings of
  the IEEE Conference on Computer Vision and Pattern Recognition Workshops.
  pp.~0--0 (2019)

\bibitem{habibie2017recurrent}
Habibie, I., Holden, D., Schwarz, J., Yearsley, J., Komura, T.: A recurrent
  variational autoencoder for human motion synthesis. In: BMVC (2017)

\bibitem{hasan2019facial}
Hasan, M.K., Rahman, W., Gerstner, L., Sen, T., Lee, S., Haut, K.G., Hoque,
  M.E.: Facial expression based imagination index and a transfer learning
  approach to detect deception  (2019)

\bibitem{lstm}
Hochreiter, S., Schmidhuber, J.: Long short-term memory. Neural computation
  \textbf{9}(8),  1735--1780 (1997)

\bibitem{hu2018deep}
Hu, G., Liu, L., Yuan, Y., Yu, Z., Hua, Y., Zhang, Z., Shen, F., Shao, L.,
  Hospedales, T., Robertson, N., et~al.: Deep multi-task learning to recognise
  subtle facial expressions of mental states. In: Proceedings of the European
  Conference on Computer Vision (ECCV). pp. 103--119 (2018)

\bibitem{JensenIdentification}
Jensen, M.L., Meservy, T.O., Kruse, J., Burgoon, J.K., Nunamaker, J.F.:
  Identification of deceptive behavioral cues extracted from video. IEEE
  Intelligent Transportation Systems pp. 1135--1140 (2005)

\bibitem{karimi2018toward}
Karimi, H., Tang, J., Li, Y.: Toward end-to-end deception detection in videos.
  In: 2018 IEEE International Conference on Big Data (Big Data). pp.
  1278--1283. IEEE (2018)

\bibitem{khodabandeh2018diy}
Khodabandeh, M., Reza Vaezi~Joze, H., Zharkov, I., Pradeep, V.: Diy human
  action dataset generation. In: Proceedings of the IEEE Conference on Computer
  Vision and Pattern Recognition Workshops. pp. 1448--1458 (2018)

\bibitem{adam}
Kingma, D.P., Ba, J.: Adam: A method for stochastic optimization. arXiv
  preprint arXiv:1412.6980  (2014)

\bibitem{kleinsmith2013affective}
Kleinsmith, A., Bianchi-Berthouze, N., et~al.: Affective body expression
  perception and recognition: A survey. IEEE TAC  (2013)

\bibitem{krishnamurthy2018deep}
Krishnamurthy, G., Majumder, N., Poria, S., Cambria, E.: A deep learning
  approach for multimodal deception detection. arXiv preprint arXiv:1803.00344
  (2018)

\bibitem{kuo2018compact}
Kuo, C.M., Lai, S.H., Sarkis, M.: A compact deep learning model for robust
  facial expression recognition. In: Proceedings of the IEEE Conference on
  Computer Vision and Pattern Recognition Workshops. pp. 2121--2129 (2018)

\bibitem{labarre1947cultural}
LaBarre, W.: The cultural basis of emotions and gestures. Journal of
  personality  \textbf{16}(1),  49--68 (1947)

\bibitem{lee2019context}
Lee, J., Kim, S., Kim, S., Park, J., Sohn, K.: Context-aware emotion
  recognition networks. In: Proceedings of the IEEE International Conference on
  Computer Vision. pp. 10143--10152 (2019)

\bibitem{LevittHawthorneEffect}
Levitt, S.D., List, J.A.: Was there really a hawthorne effect at the hawthorne
  plant? an analysis of the original illumination experiments. American
  Economic Journal: Applied Economics  \textbf{3}(1),  224--38 (2011)

\bibitem{marinoiu20183d}
Marinoiu, E., Zanfir, M., Olaru, V., Sminchisescu, C.: 3d human sensing, action
  and emotion recognition in robot assisted therapy of children with autism.
  In: Proceedings of the IEEE Conference on Computer Vision and Pattern
  Recognition. pp. 2158--2167 (2018)

\bibitem{MasipDefiningDeception}
Masip, J., Garrido, E., Herrero, C.: Defining deception. Anales de Psicolog\'ia
   \textbf{20}(1),  147--171 (2004)

\bibitem{mcduff2017large}
McDuff, D., Soleymani, M.: Large-scale affective content analysis: Combining
  media content features and facial reactions. In: 2017 12th IEEE International
  Conference on Automatic Face \& Gesture Recognition (FG 2017). pp. 339--345.
  IEEE (2017)

\bibitem{meservy2005deception}
Meservy, T.O., Jensen, M.L., Kruse, J., Burgoon, J.K., Nunamaker, J.F.,
  Twitchell, D.P., Tsechpenakis, G., Metaxas, D.N.: Deception detection through
  automatic, unobtrusive analysis of nonverbal behavior. IEEE Intelligent
  Systems  \textbf{20}(5),  36--43 (2005)

\bibitem{michael2010motion}
Michael, N., Dilsizian, M., Metaxas, D., Burgoon, J.K.: Motion profiles for
  deception detection using visual cues. In: European Conference on Computer
  Vision. pp. 462--475. Springer (2010)

\bibitem{michalak2009embodiment}
Michalak, J., Troje, N.F., Fischer, J., Vollmar, P., Heidenreich, T., Schulte,
  D.: Embodiment of sadness and depression—gait patterns associated with
  dysphoric mood. Psychosomatic medicine  \textbf{71}(5),  580--587 (2009)

\bibitem{milgram1978obedience}
Milgram, S., Gudehus, C.: Obedience to authority (1978)

\bibitem{morais2019learning}
Morais, R., Le, V., Tran, T., Saha, B., Mansour, M., Venkatesh, S.: Learning
  regularity in skeleton trajectories for anomaly detection in videos. In:
  Proceedings of the IEEE Conference on Computer Vision and Pattern
  Recognition. pp. 11996--12004 (2019)

\bibitem{mount2008intentions}
Mount, A.: Intentions, gestures, and salience in ordinary and deferred
  demonstrative reference. Mind \& Language  \textbf{23}(2),  145--164 (2008)

\bibitem{Navarro2003}
Navarro, J.: Four-domain model for detection deception: An alternative paradigm
  for interviewing. FBI Law Enforcement Bulletin  \textbf{72}(6),  19--24
  (2003)

\bibitem{nguyen2019anomaly}
Nguyen, T.N., Meunier, J.: Anomaly detection in video sequence with
  appearance-motion correspondence. In: Proceedings of the IEEE International
  Conference on Computer Vision. pp. 1273--1283 (2019)

\bibitem{noroozi2018survey}
Noroozi, F., Kaminska, D., Corneanu, C., Sapinski, T., Escalera, S.,
  Anbarjafari, G.: Survey on emotional body gesture recognition. IEEE
  transactions on affective computing  (2018)

\bibitem{pavllo2018quaternet}
Pavllo, D., Grangier, D., Auli, M.: Quaternet: A quaternion-based recurrent
  model for human motion. arXiv preprint arXiv:1805.06485  (2018)

\bibitem{perez2015deception}
P{\'e}rez-Rosas, V., Abouelenien, M., Mihalcea, R., Burzo, M.: Deception
  detection using real-life trial data. In: Proceedings of the 2015 ACM on
  International Conference on Multimodal Interaction. pp. 59--66. ACM (2015)

\bibitem{perez2015verbal}
P{\'e}rez-Rosas, V., Abouelenien, M., Mihalcea, R., Xiao, Y., Linton, C.,
  Burzo, M.: Verbal and nonverbal clues for real-life deception detection. In:
  Proceedings of the 2015 Conference on Empirical Methods in Natural Language
  Processing. pp. 2336--2346 (2015)

\bibitem{randhavane2019identifying}
Randhavane, T., Bera, A., Kapsaskis, K., Bhattacharya, U., Gray, K., Manocha,
  D.: Identifying emotions from walking using affective and deep features.
  arXiv preprint arXiv:1906.11884  (2019)

\bibitem{rill2019high}
Rill-Garcia, R., Jair~Escalante, H., Villasenor-Pineda, L., Reyes-Meza, V.:
  High-level features for multimodal deception detection in videos. In:
  Proceedings of the IEEE Conference on Computer Vision and Pattern Recognition
  Workshops. pp.~0--0 (2019)

\bibitem{roether2009critical}
Roether, C., Omlor, L., et~al.: Critical features for the perception of emotion
  from gait. Vision  (2009)

\bibitem{shi2019skeleton}
Shi, L., et~al.: Skeleton-based action recognition with directed graph neural
  networks. In: Proceedings of the IEEE Conference on Computer Vision and
  Pattern Recognition (2019)

\bibitem{Slepian}
Slepian, M.L., Masicampo, E.J., Toosi, N.R., Ambady, N.: The physical burdens
  of secrecy. Journal of Experimental Psychology: General  \textbf{141}(4),
  619--624 (2012)

\bibitem{sultani2018real}
Sultani, W., Chen, C., Shah, M.: Real-world anomaly detection in surveillance
  videos. In: Proceedings of the IEEE Conference on Computer Vision and Pattern
  Recognition. pp. 6479--6488 (2018)

\bibitem{tsikerdekis2014online}
Tsikerdekis, M., Zeadally, S.: Online deception in social media. Communications
  of the ACM  \textbf{57}(9), ~72 (2014)

\bibitem{umeyama}
Umeyama, S.: Least-squares estimation of transformation parameters between two
  point patterns. TPAMI (4),  376--380 (1991)

\bibitem{vrij2019reading}
Vrij, A., Hartwig, M., Granhag, P.A.: Reading lies: nonverbal communication and
  deception. Annual review of psychology  \textbf{70},  295--317 (2019)

\bibitem{wan2019survey}
Wan, C., Wang, L., Phoha, V.V.: A survey on gait recognition. ACM Computing
  Surveys (CSUR)  \textbf{51}(5), ~89 (2019)

\bibitem{wang2015video}
Wang, S., Ji, Q.: Video affective content analysis: a survey of
  state-of-the-art methods. IEEE Transactions on Affective Computing
  \textbf{6}(4),  410--430 (2015)

\bibitem{wang2019ev}
Wang, Y., Du, B., Shen, Y., Wu, K., Zhao, G., Sun, J., Wen, H.: Ev-gait:
  Event-based robust gait recognition using dynamic vision sensors. In:
  Proceedings of the IEEE Conference on Computer Vision and Pattern
  Recognition. pp. 6358--6367 (2019)

\bibitem{wei2017deep}
Wei, X.S., Zhang, C.L., Zhang, H., Wu, J.: Deep bimodal regression of apparent
  personality traits from short video sequences. IEEE Transactions on Affective
  Computing  \textbf{9}(3),  303--315 (2017)

\bibitem{wu2017recent}
Wu, D., Sharma, N., Blumenstein, M.: Recent advances in video-based human
  action recognition using deep learning: A review. In: 2017 International
  Joint Conference on Neural Networks (IJCNN). pp. 2865--2872. IEEE (2017)

\bibitem{wu2018deception}
Wu, Z., Singh, B., Davis, L.S., Subrahmanian, V.: Deception detection in
  videos. In: Thirty-Second AAAI Conference on Artificial Intelligence (2018)

\bibitem{xu2016heterogeneous}
Xu, B., Fu, Y., Jiang, Y.G., Li, B., Sigal, L.: Heterogeneous knowledge
  transfer in video emotion recognition, attribution and summarization. IEEE
  Transactions on Affective Computing  \textbf{9}(2),  255--270 (2016)

\bibitem{yan2018spatial}
Yan, S., Xiong, Y., Lin, D.: Spatial temporal graph convolutional networks for
  skeleton-based action recognition. In: Thirty-Second AAAI Conference on
  Artificial Intelligence (2018)

\bibitem{yang2018pose}
Yang, C., Wang, Z., Zhu, X., Huang, C., Shi, J., Lin, D.: Pose guided human
  video generation. In: Proceedings of the European Conference on Computer
  Vision (ECCV). pp. 201--216 (2018)

\bibitem{yao2019review}
Yao, G., Lei, T., Zhong, J.: A review of convolutional-neural-network-based
  action recognition. Pattern Recognition Letters  \textbf{118},  14--22 (2019)

\bibitem{yu2009study}
Yu, S., Tan, T., Huang, K., Jia, K., Wu, X.: A study on gait-based gender
  classification. IEEE Transactions on image processing  \textbf{18}(8),
  1905--1910 (2009)

\bibitem{van2019freeze}
Van~der Zee, S., Poppe, R., Taylor, P.J., Anderson, R.: To freeze or not to
  freeze: A culture-sensitive motion capture approach to detecting deceit. PloS
  one  \textbf{14}(4),  e0215000 (2019)

\bibitem{zhang2016deep}
Zhang, C.L., Zhang, H., Wei, X.S., Wu, J.: Deep bimodal regression for apparent
  personality analysis. In: European Conference on Computer Vision. pp.
  311--324. Springer (2016)

\bibitem{zhang2019learning}
Zhang, K., Luo, W., Ma, L., Liu, W., Li, H.: Learning joint gait representation
  via quintuplet loss minimization. In: Proceedings of the IEEE Conference on
  Computer Vision and Pattern Recognition. pp. 4700--4709 (2019)

\bibitem{zhang2019gait}
Zhang, Z., Tran, L., Yin, X., Atoum, Y., Liu, X., Wan, J., Wang, N.: Gait
  recognition via disentangled representation learning. In: Proceedings of the
  IEEE Conference on Computer Vision and Pattern Recognition. pp. 4710--4719
  (2019)

\bibitem{zhou2008following}
Zhou, L., Zhang, D.: Following linguistic footprints: automatic deception
  detection in online communication. Commun. ACM  \textbf{51}(9),  119--122
  (2008)

\bibitem{Zuckerman1981}
Zuckerman, M., DePaulo, B., Rosenthal, R.: Verbal and nonverbal communication
  of deception. Advances in Experimental Social Psychology  \textbf{14},  1--59
  (1981)

\bibitem{zuo2019your}
Zuo, J., Gedeon, T., Qin, Z.: Your eyes say you’re lying: An eye movement
  pattern analysis for face familiarity and deceptive cognition. In: 2019
  International Joint Conference on Neural Networks (IJCNN). pp.~1--8. IEEE
  (2019)

\end{thebibliography}
